\documentclass{article}

\usepackage[dblblindworkshop, final]{neurips_2025}

\usepackage[utf8]{inputenc} % allow utf-8 input
\usepackage[T1]{fontenc}    % use 8-bit T1 fonts
\usepackage{hyperref}       % hyperlinks
\usepackage{url}            % simple URL typesetting
\usepackage{booktabs}       % professional-quality tables
\usepackage{amsfonts}       % blackboard math symbols
\usepackage{nicefrac}       % compact symbols for 1/2, etc.
\usepackage{microtype}      % microtypography
\usepackage[dvipsnames]{xcolor}
\usepackage{amsmath}
\usepackage{graphicx}

\usepackage[capitalize,nameinlink]{cleveref}

\newcommand{\xhdr}[1]{\textbf{#1}\:}

\title{Predicting symbolic ODEs from multiple trajectories}
\workshoptitle{Machine Learning and the Physical Sciences}

\author{%
  Yakup Emre Şahin\phantom{mmm} Niki Kilbertus\phantom{mmm}  Sören Becker\\
  \phantom{mmm}Helmholtz Munich\\
  \phantom{mmm}Technical University of Munich\\
  \phantom{mmm}Munich Center for Machine Learning (MCML) \\
  \phantom{mmm}\texttt{first.last@helmholtz-munich.de} \\
}

\begin{document}

\bibliographystyle{plainnat}

\maketitle

\begin{abstract}
  We introduce MIO, a transformer-based model for inferring symbolic ordinary differential equations (ODEs) from multiple observed trajectories of a dynamical system. By combining multiple instance learning with transformer-based symbolic regression, the model effectively leverages repeated observations of the same system to learn more generalizable representations of the underlying dynamics. We investigate different instance aggregation strategies and show that even simple mean aggregation can substantially boost performance. MIO is evaluated on systems ranging from one to four dimensions and under varying noise levels, consistently outperforming existing baselines.
\end{abstract}

\section{Introduction}

Identifying governing dynamics equations from data is a core objective in scientific modeling as it enables not only accurate predictions, but also provides understanding of the underlying system structure and thus ultimately scientific insight. Symbolic regression directly supports this goal by recovering closed-form, interpretable mathematical expressions that describe a system's behavior. While classical symbolic regression methods perform explicit search over symbolic expression spaces using techniques like genetic programming \citep{koza, schmidt2010age, cranmer2023interpretablemachinelearningscience}, tree-based search \citep{jin2019bayesian}, probabilistic grammars \citep{brence2021probabilistic}, or basis function regression \citep{McConaghy2011, brunton_discovering_2016}, recent neural approaches learn this mapping during pretraining and hence translate input data into symbolic expressions more efficiently at inference time \citep{nesymres, valipour2021symbolicgpt, vastl2022symformer, kamienny2022end, meidani_snip_2023}.
While transformer-based models were originally not developed for dynamics equations,
they have been extended to infer both univariate \citep{becker2023predicting} and systems of ordinary differential equations (ODEs) \citep{d'ascoli2024odeformer}, offering improved robustness to noise and irregular sampling compared to classical methods by avoiding numerical differentiation. 

A key property of differential equations is that they capture the dynamics irrespective of the particular state a system is in.
Accordingly, inferred equations should generalize to unseen system states. However, accurate identification from a single trajectory can be difficult or even fundamentally impossible due to noise, sparsity \citep{casolo2025identifiability}, or structural ambiguity, even in linear systems \citep{stanhope2014identifiability} and especially in nonlinear ones \citep{scholl2023uniqueness}.
In practice though, multiple trajectories of the same system are often available, e.g., in repeated measures designs, and can intuitively aid identification. Yet, existing transformer-based models are limited to processing one trajectory at a time.
We address this limitation with \textbf{Multiple Instance-ODEFormer (MIO)}, a model that applies multiple instance learning \citep{ilse2018attention, chen2024timemil} to infer symbolic ordinary differential equations from multiple observations, thereby leveraging the full informational richness of the dataset for more accurate system identification. Code for our project is available at \texttt{\url{https://github.com/yakupemresahin/mio}}.

\section{Methods} \label{sec:methods}

MIO follows the well-established sequence-to-sequence modeling paradigm \citep{vaswani2023attentionneed} and maps observed input trajectories to symbolic ODE expressions. It is trained on a large corpus of synthetically generated ODEs after which predictions on new data require no additional fitting time.

\xhdr{Data generation.} \label{sec:data_generation}
To generate the training corpus, we follow the procedure presented by \citet{d'ascoli2024odeformer} with hyperparameter choices listed in \cref{app:sec:data_gen_details}. This procedure first generates a mathematical expression that is interpreted as the right-hand side of an ODE and which is solved numerically, thus producing paired data of numerical trajectories and corresponding symbolic ODE expressions.

\begin{figure}
    \centering
    \includegraphics[width=1\linewidth, trim=8 12 10 9, clip]{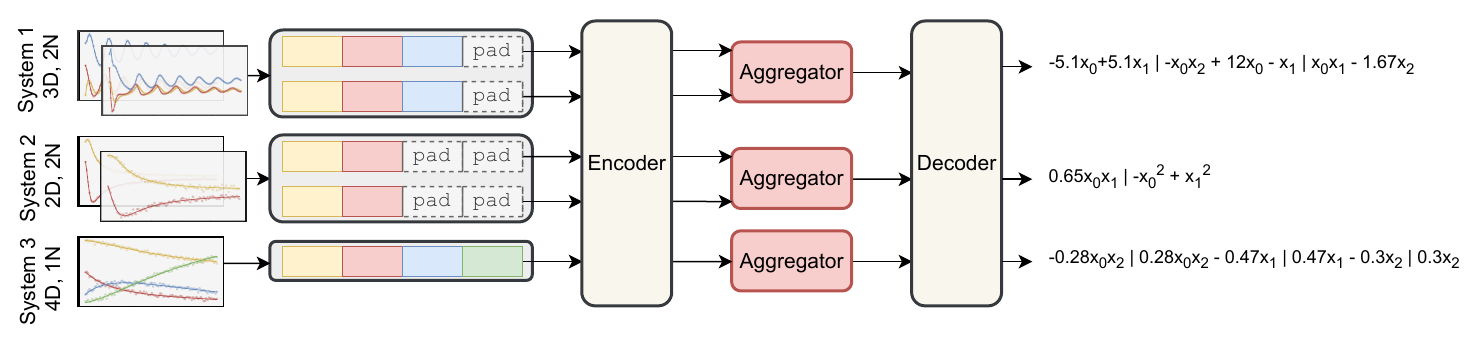}
    \caption{\textbf{Model overview.} System dimensions (2D, 3D, 4D) and \# instances (2N, 2N, 1N) may vary.}
    \label{fig:overview}
\end{figure}

\xhdr{Model architecture.} MIO builds directly on ODEFormer \citep{d'ascoli2024odeformer} and keeps the trajectory embedder, encoder and decoder unchanged (details in \cref{app:sec:training_details}), while introducing an additional aggregator block between encoder and decoder as depicted in \cref{fig:overview}. The encoder $\mathcal{E}$ processes each trajectory instance separately, similar to mini-batch elements, and produces instance-specific latents $\mathbf{z}_j = \mathcal{E}(\mathbf{h}_j) \in \mathbb{R}^{s\times d_{\text{emb}}}$ from embedded trajectories $\mathbf{h}_j$, where $s$ denotes the number of points in the trajectory, $d_{\text{emb}}$ denotes the embedding dimension and $j$ indexes the input trajectories. We keep track of the number of instances per system so that only instances of the same underlying ODE are combined by the aggregator. The aggregated system-specific latents $\mathbf{\bar z}$ are passed to the decoder, which produces a single prediction per system. The architecture can handle variable numbers of instances per system and inherits ODEFormer's flexibility to handle different system dimensionalities.

\xhdr{Aggregation strategies.} The aggregator is the central innovation of our architecture, designed to fuse the information of all available instance embeddings $\{ \mathbf{z}_1, ..., \mathbf{z}_n \}$ into a single system representation $\mathbf{\bar z}$. We explore several aggregation strategies, the first of which is \textbf{mean pooling} which simply averages instance embeddings $ \mathbf{\bar z} = \frac{1}{n} \sum_{j=1}^n \mathbf{z}_j \in \mathbb{R}^{s \times d_{\text{emb}}}$. Mean pooling is parameter-free, making it fast and memory-efficient. However, assuming all instances contribute equally to system identification can be limiting so that the computational benefit may come at the cost of sub-optimal performance. 

As a second aggregation strategy we assess \textbf{attentive pooling}. In this case we first perform aggregation over time to incorporate temporal structure more explicitly similar to \citet{meidani_snip_2023}. Specifically, we pass each instance-specific latent $\mathbf{z}_j \in \mathbb{R}^{s \times d_\text{emb}}$ through a 4-layer transformer encoder $\mathcal{A}_\text{time}$ and use the class-token embedding of the final layer as condensed instance representation $\mathbf{\tilde{z}}_j \in \mathbb{R}^{d_{\text{emb}}}$. Subsequently, instance representations are combined by weighted averaging $\mathbf{\bar z} = \sum_{j=1}^n \omega_j \cdot \mathbf{z}_j \in \mathbb{R}^{s \times d_{\text{emb}}}$ where weights are defined by a softmax over condensed instances $\omega_j = \nicefrac{\exp(\mathbf{w^T} \mathbf{\tilde{z}}_j)}{\sum_{j=1}^n \exp(\mathbf{w^T} \mathbf{\tilde{z}}_j})$ with learnable parameters $\mathbf{w} \in \mathbb{R}^{d_{\text{emb}}}$. Attentive pooling essentially weighs different instances according to their relevance in comparison to other instances. 

Alternatively, we replace (weighted) averaging altogether by aggregating via attention. As a first option, we propose \textbf{time-agnostic attention pooling}. Assume the input to be the tensor $\mathbf{Z} \in \mathbb{R}^{s \times n \times d_{\text{emb}}}$, where $s, n, d_{\text{emb}}$ correspond to the number of points in the trajectory, the number of instances and the embedding dimension. We use this representation directly as keys and values in a cross-attention layer where the query is a learnable parameter $\mathbf{q}_\text{a} \in \mathbb{R}^{d_{\text{emb}}}$ which we expand to shape $\mathbb{R}^{s\times1\times d_{\text{emb}}}$ using \texttt{torch.expand()}. 
This aggregation method attends to all instances at once in a time-resolved manner and, in contrast to the softmax operation in attentive pooling, does not highlight the relevance of any single instance at the expense of other instances. However, this aggregation ignores the temporal structure of the input and processes each embedding of the input sequence independently.

As a remedy we finally introduce \textbf{time-aware attention pooling}. As with time-agnostic attention pooling, we start with a tensor $\mathbf{Z} \in \mathbb{R}^{s \times n \times d_{\text{emb}}}$ which we concatenate with the time-aggregated representations $\mathbf{\tilde{z}}_j$ as well as a class token embedding to form the input tensor $\mathbf{Z'} \in \mathbb{R}^{s \times (2n + 1) \times d_{\text{emb}}}$ where the $2n+1$ dimensions correspond to $n$ trajectory embeddings $\mathbf{z}_j \in \mathbb{R}^{s \times d_{\text{emb}}}$, $n$ condensed embeddings $\mathbf{\tilde{z}}_j \in \mathbb{R}^{d_{\text{emb}}}$ and a learnable class token embedding $\mathbf{c} \in \mathbb{R}^{d_{\text{emb}}}$, where every $\mathbf{\tilde{z}}_j$ and $\mathbf{c}$ is expanded to match the shape of $\mathbf{z}_j$. We use a 4-layer transformer encoder to process $\mathbf{Z'}$ and use the resulting output embedding of the class token $\mathbf{\bar{z}} \in \mathbb{R}^{s \times d_{\text{emb}}}$ as system representation.
As in time-agnostic attention pooling, time-aware attention pooling processes embeddings in the input sequence separately, however, temporal information is available to the aggregation method as we explicitly add representations $\mathbf{\tilde{z}}_j$ as input.

\section{Results}

\begin{figure}[t]
    \centering
    \includegraphics[width=0.49\linewidth]{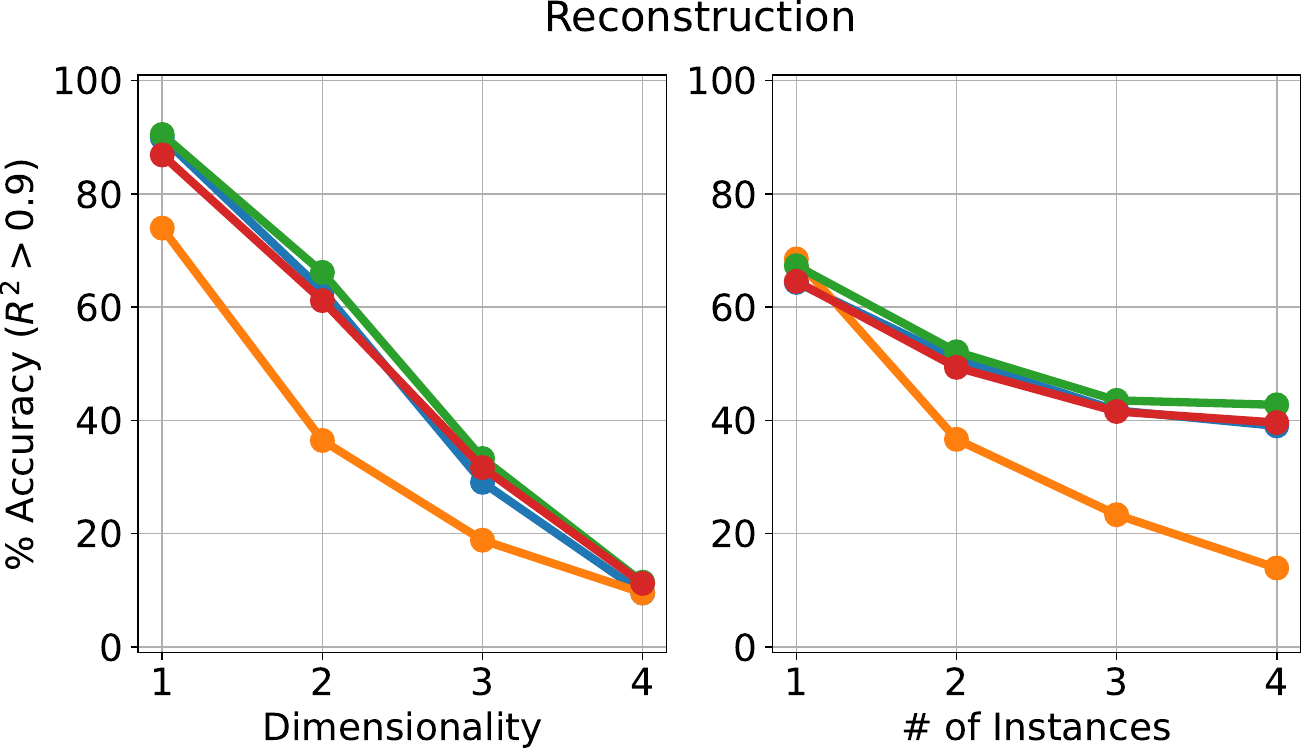}
    \hspace{0.5pt}
    \includegraphics[width=0.49\linewidth]{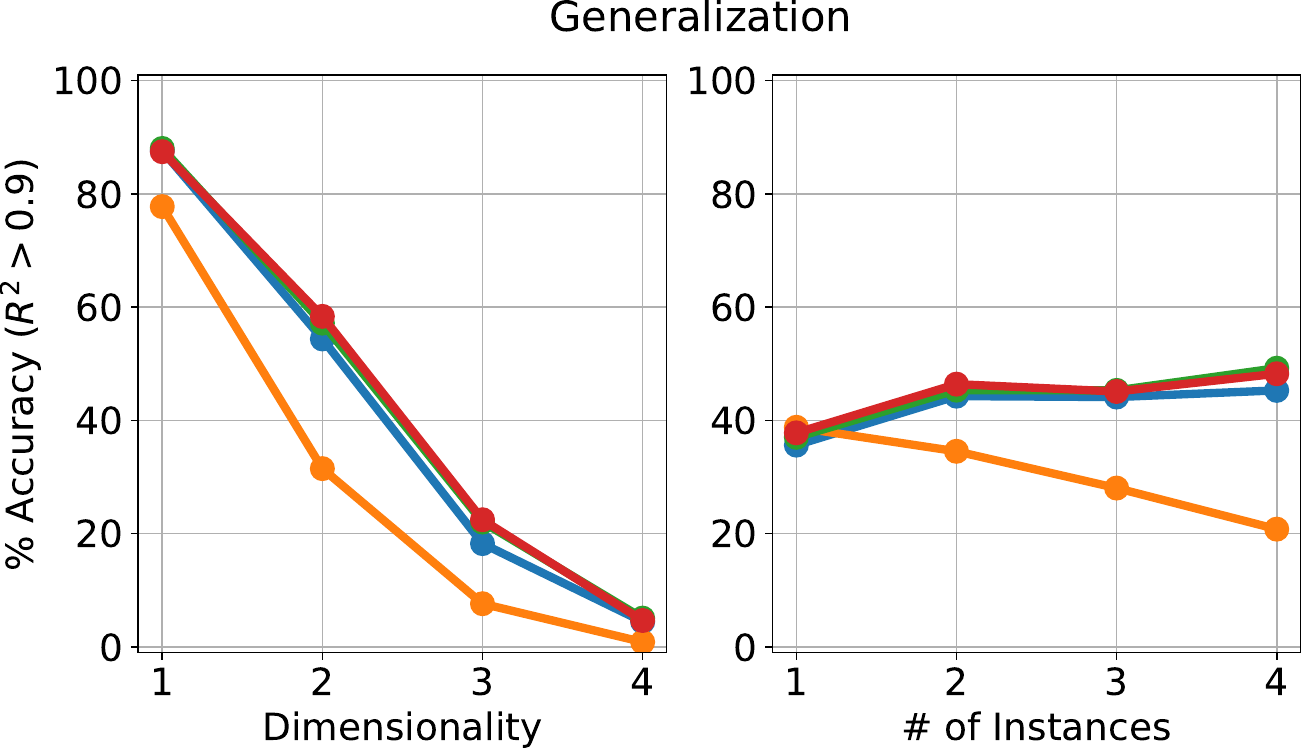}
    \includegraphics[width=0.75\linewidth]{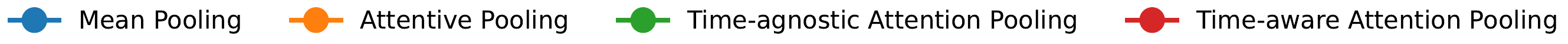}
    \caption{Performance comparison of different instance aggregation methods.}
    \label{fig:secondary}
\end{figure}

\xhdr{Evaluation tasks.} 
Following \citet{d'ascoli2024odeformer}, we evaluate model performance on two tasks: reconstruction and generalization. 
Reconstruction assesses if the trajectories obtained by solving the models’ predicted equations align with the observed input trajectories. In contrast, the more challenging generalization task assesses the alignment for previously unseen trajectories, i.e., numerical solutions to the ground truth and predicted ODE for a new initial value that was not used to infer the symbolic form of the ODE. Performance in both tasks is evaluated in terms of accuracy, which, following \citet{d'ascoli2024odeformer}, is defined as the fraction of predicions for which the $R^2$ score between (noiseless) ground truth and predicted trajectories exceeds a threshold of $0.9$. In case of reconstruction, we obtain a separate $R^2$ score for each observed instance; we assess if the minimum of these exceeds the threshold as we seek a single ODE to model all instances.

\xhdr{Experiment 1: How do different aggregation methods compare?}
We train a single model per aggregation method on $\sim$25M ODE systems whose dimensions vary between one and four and which each come with up to four instances. Our aim in this pilot experiment is to compare the aggregation methods. We therefore settle for a simplified dataset which only contains polynomials and initialize the weights with the original ODEFormer weights to speed up training. The test set contains 2000 systems, roughly 500 per dimension, with approximately 500 systems for 1, 2, 3 and 4 instances. Details on data and model are provided in \cref{app:sec:data_gen_details,app:sec:training_details,}.

As presented in \cref{fig:secondary}, mean pooling, time-aware attention pooling and time-agnostic attention pooling perform on par and are clearly superior to attentive pooling. Surprisingly, simple mean aggregation performs only marginally worse than the best attention-based pooling schemes across all tested number of dimensions and instances. An additional interesting trend is that an increasing number of instances improves performance in case of the generalization task yet hurts reconstruction performance. This is because generalization requires fitting a single, unseen trajectory, regardless of how many instances inform the prediction. In contrast, reconstruction requires the predicted ODE to fit all input instances, so more instances do not necessarily reduce task difficulty. Finally, a striking trend is the performance degradation as the system dimensionality increases. A potential explanation for this behavior is that the number of training samples required for good performance increases with system dimension whereas it is roughly equal in our training dataset.

\xhdr{Experiment 2: Comparison with baselines and across noise levels.}
Based on the results of the initial experiment, we use mean pooling due to its computational efficiency at negligible performance loss. Moreover, we focus on 2D and 3D systems as our initial results indicate that these are far more challenging than 1D systems yet not as far out of reach as 4D systems. At the same time, we drop the polynomial data restriction to include more general non-linear systems and corrupt the trajectories with multiplicative Gaussian noise sampled independently per time step from $\mathcal{N}(1, \sigma^2)$ with $\sigma = 0.05$. The final training datasets contain $\sim$45M 2D systems and $\sim$55M 3D systems.

\begin{figure}[]
    \centering
    \includegraphics[width=0.49\linewidth]{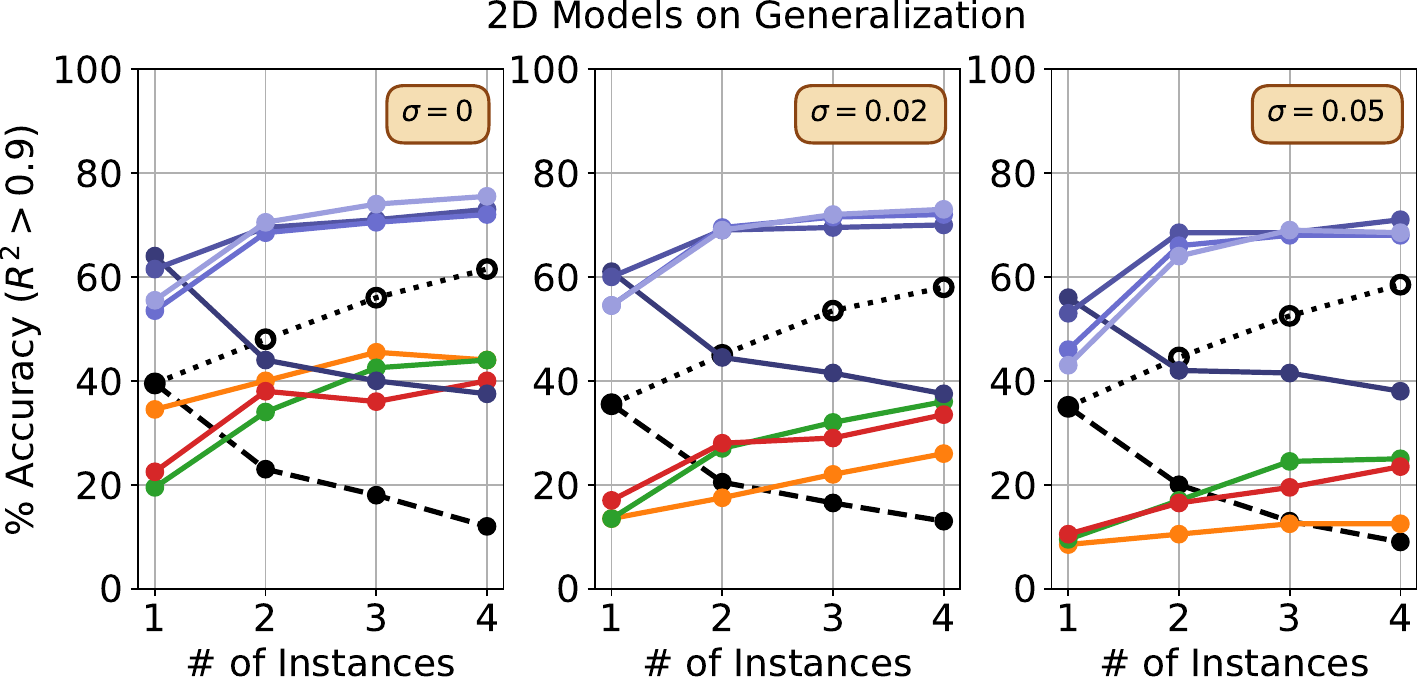}
    \hspace{0.5pt}
    \includegraphics[width=0.49\linewidth]{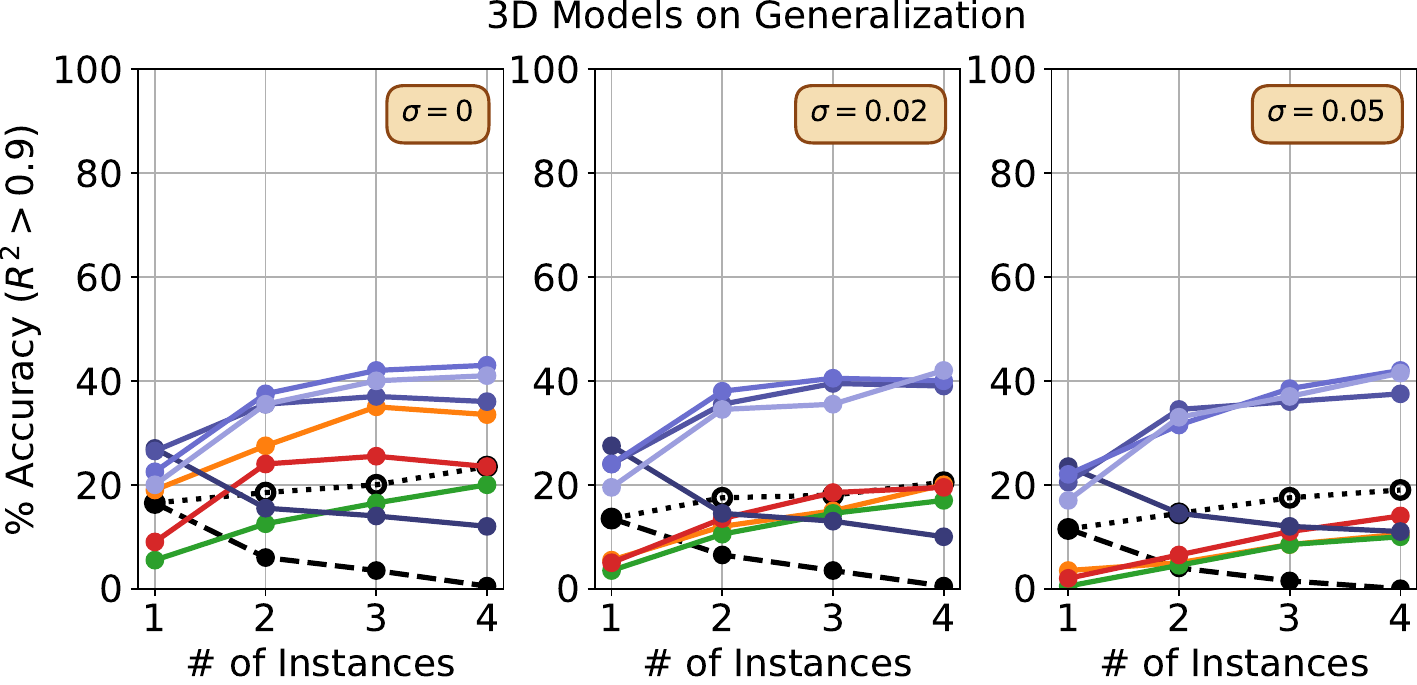}
    \includegraphics[width=\linewidth]{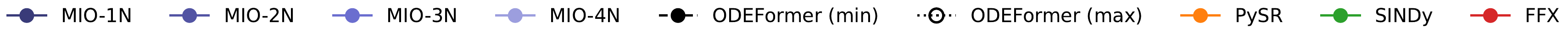}
    \caption{Performance comparison across system dimensions, number of instances and noise levels.}
    \label{fig:pilot}
\end{figure}

We train separate models per number of instances on 2D and 3D systems (8 models in total; 4 instances $\times$ 2 dimension); additional training details are provided in \cref{app:sec:training_details}. We compare the performance of our models to four baseline methods: PySR \citep{cranmer2023interpretablemachinelearningscience}, which is based on evolutionary algorithms, SINDy \citep{brunton_discovering_2016} and FFX \citep{McConaghy2011}, which are regression based methods with fixed basis functions, as well as ODEFormer \citep{d'ascoli2024odeformer} which follows the transformer-based sequence-to-sequence modeling paradigm. While ODEFormer builds the basis for our model, it is fundamentally unable to process multiple instances. As a workaround we run ODEFormer on individual instances to obtain multiple predictions per system and use the best (ODEFormer (max)) and worst (ODEFormer (min)) fitting ODE to compute performances. Note that using the best predicted ODE clearly favors ODEFormer in the comparison as it corresponds to a top-n evaluation whereas all other models are evaluated in a top-1 fashion. In particular, this gives ODEFormer the possibility to focus on the single observed trajectory that is closest to the test set trajectory in the generalization case whereas this information is not available to the rest of the models. Additional information on the evaluation of baselines is provided in \cref{app:sec:baseline_configs}.

Performance is evaluated on 200 systems in both 2D and 3D that are generated as described in \cref{sec:data_generation}. We focus on the generalization task here and report results for reconstruction in \cref{app:sec:additional_results}. As depicted in \cref{fig:pilot}, performance increases substantially with number of instances if MIO is trained on more than one instance (2N-4N models). Interestingly, performances between models within this group differ only marginally.
This is in stark contrast to models trained on a single instance only (1N models), for which additional instances degrade performance. While absolute performances decrease with increasing dimensionality, they are robust across noise levels. Even though all baselines are clearly outperformed by our model, PySR, SINDy and FFX reflect the trends, except that their performances suffer as the noise level $\sigma$ increases. For ODEFormer (max) we see a linear performance increase that reflects its custom top-n evaluation scheme. Interestingly, when moving from one to two instances, MIO shows a larger performance gain than ODEFormer (max), indicating that our model leverages the additional information beyond what a top-2 evaluation would allow for.

\section{Conclusion}
In this work, we studied how to aggregate multiple observed trajectories for symbolic ODE discovery, a core challenge in learning dynamical models from limited and noisy data, and found that, surprisingly, simple mean pooling performs on par with more sophisticated alternatives. As long as the model is trained on multiple instances, performance improves as expected with the number of available instances at test time, with the largest gain observed when moving from one to two instances. While our current evaluation is limited to test sets drawn from the same distribution as the training data, our focus is on understanding aggregation strategies rather than building a data distribution agnostic model or a model for a particular application domain. Within this controlled setting, MIO outperforms both equation-specific baselines and ODEFormer which was optimized on a similar training distribution as our model. As future work, we aim to unify our dimension- and instance-specific models into a single, generalizable one.

Our results open up exciting research questions, for example why the improvements in performance diminish after observing more than two instances. A potential hypothesis here is that the dimensionality of the tested systems is too low so that the complexity of their behavior is too limited and does in many cases not require more than two trajectories. This hypothesis is consistent with the observation that mean pooling performs well in comparison to other strategies: in low dimensions the observational space might still be sufficiently small so that mean pooling can capture the observable variation in systems. That is, since the model already applies mean pooling during training, the space between instance-specific latents might already be well traversed during training as low dimensional systems (especially 1D and 2D systems) are limited in the qualitative complexity of their behavior. Lastly, a highly relevant open challenge that this work did not yet touch upon is the scalability to higher dimensions using this modeling approach to symbolic regression.

\section{Acknowledgements}
This work has been supported by the German Federal Ministry of Education and Research (Grant: 01IS24082). The authors also gratefully acknowledge the scientific support and resources of the AI service infrastructure LRZ AI Systems provided by the Leibniz Supercomputing Centre (LRZ) of the Bavarian Academy of Sciences and Humanities (BAdW), funded by Bayerisches Staatsministerium für Wissenschaft und Kunst (StMWK). SB is supported by the Helmholtz Association under the
joint research school “Munich School for Data Science - MUDS”.

\bibliography{ref}

\clearpage
\appendix

\section{Details of data generation}
\label{app:sec:data_gen_details}

In this section, we describe how the data were generated for experiments 1 and 2. While the construction of symbolic ODEs differs between the two experiments, the subsequent numerical procedure to solve them is identical: after obtaining the mathematical expression, we solve every ODE on the time interval $[1, 10 ]$ with 100 equidistant support points using \texttt{scipy.integrate.odeint} with relative and absolute tolerances set to $\texttt{rtol}=10^{-3}$ and $\texttt{atol}= 10^{-6}$. For each system, we sample four initial values from a standard normal distribution. Numerical solutions in which any component exceeds an absolute value of $10^2$ are discarded, serving as an amplitude filter to prevent an over-abbundance of diverging systems.

In experiment 1, both the dimensionality of the system and the number of instances per system are varied, each sampled uniformly from $[1, 4]$. In contrast, experiment 2 focuses on assessing the effect of additional instances on performance. Here, for each 2D and 3D system, we generate four instances in total and construct the $n$-instance model by selecting the first $n$ instances from these. This ensures that the data distribution remains fixed while allowing models with different numbers of instances to span the same underlying systems.

\xhdr{ODE generation in experiment 1.}
We generate polynomial ODE systems by first sampling a maximum polynomial order $o_\text{max}$ uniformly from the range $[1, O_\text{max}=3]$. Together with the dimensionality $D$, this parameter determines the structural complexity of the dynamical system.

Given a particular choice of $D$ and $o_\text{max}$, we enumerate all valid monomial terms that can appear in the symbolic expressions of the time derivatives $\dot{x}_i$. Specifically, we collect all exponent vectors $\mathbf{o} \in \mathbb{N}^D$ such that the total degree satisfies $|\mathbf{o}| \leq o_\text{max}$. Each such vector defines a monomial of the form $x_1^{o_1} x_2^{o_2} \cdots x_D^{o_D}$. For example, in a two-dimensional system ($D=2$) with $o_\text{max}=3$, the valid exponent vectors are
$\mathbf{o} = \{(0,0), (0,1), (0,2), (0,3), (1,0), (1,1), (1,2), (2,0), (2,1), (3,0)\}$,
which span the full basis of candidate terms for each equation $\dot{x}_i$.

For every monomial, we assign a coefficient sampled from a log-normal distribution with parameters $\mu=0$ and $\sigma=1$, ensuring variability across multiple orders of magnitude. The sign of each coefficient is chosen uniformly at random from $\{+,-\}$, and the final value is rounded to five decimal places for consistency in symbolic representation.

Next, for each component equation $\dot{x}_i = f_i(\mathbf{x})$, we sample the number of terms $n_\text{terms}$ from a truncated normal distribution with $\mu=2$ and $\sigma=2$, trimmed to the range $[1, N^\text{max}_\text{terms} = 5]$. After rounding to the nearest integer, if the number of candidate terms exceeds $n_\text{terms}$, we randomly subsample to match this value. This procedure ensures that each ODE component is expressed as a concise yet sufficiently expressive polynomial, balancing training efficiency and symbolic interpretability.

\xhdr{ODE generation in experiment 2.}
For Experiment 2, we generate two- and three-dimensional ODE systems following the data generation scheme of \citet{d'ascoli2024odeformer}, with minor modifications. Unary operators are sampled from the set $\{\sin(x), x^2, x^{-1}, \mathrm{id}(x)\}$ with probabilities $\nicefrac{1}{6}, \nicefrac{1}{6}, \nicefrac{1}{6}, \nicefrac{1}{2}$, respectively. Binary operators are chosen with probabilities $P(+)=\nicefrac{3}{4}$ and $P(\times)=\nicefrac{1}{4}$. The full set of hyperparameters used for tree generation is provided in \cref{tab:exp2_data_hyperparameters}.

\begin{table}[h!]
  \caption{Hyperparameters of ODE generation for Experiment 2}
  \label{tab:exp2_data_hyperparameters}
  \centering
  \begin{tabular}{lll}
    \toprule
    Hyperparameter     & Value     & Description \\
    \midrule
    $b_\text{max}$ & 5 & maximum number of binary operators\\
    $u_\text{max}$ & 3 & maximum number of unary operators\\
    $(c_\text{min}, c_\text{max})$ & (0.05, 20) & parameters of log-uniform distribution for affine transformation\\
    $d_\text{max}$ & 6 & maximum depth of subtrees \\
    \bottomrule
  \end{tabular}
\end{table}

\section{Details on MIO: architecture, training, inference} \label{app:sec:training_details}

\subsection{Model architecture}
MIO mirrors the base architecture components of ODEFormer \citep{d'ascoli2024odeformer}. In particular the model follows the classical embedder-encoder-decoder transformer design \citep{vaswani2023attentionneed}. 

To represent floating-point values found in both numeric input trajectories and symbolic target sequences, we require an efficient encoding scheme that balances precision with a fixed vocabulary size. Following \citet{d'ascoli2024odeformer}, we tokenize each float by:
\begin{enumerate}
    \item Rounding it to four significant digits.
    \item Decomposing it into three components: sign, mantissa, and exponent.
    \item Encoding each component as a separate token.
\end{enumerate}

This three-token scheme reduces the vocabulary size needed to represent floating-point numbers to just 10,203 tokens (including $+$, $-$, 0-9999, and exponent terms E-100 to E100). Despite the precision loss, this method performs well in practice.
In a D-dimensional ODE system, each observed time point $(t_i, \mathbf{x}_i)$ in the input trajectory is tokenized using the above scheme, resulting in three tokens per time point.
Each such token is independently embedded, and the resulting embeddings are concatenated into a vector of shape $3(D+1) \times d_\text{emb}$ for that time point. To support variable input dimensions (up to a maximum $D_{\text{max}}=4$), systems with $D<D_{\text{max}}$ are zero-padded. This concatenated embedding is then passed through a 2-layer feedforward network, reducing it to a final representation of dimension $d_\text{emb}$ per time point. We use $h_i \in \mathbb{R}^{s \times d_\text{emb}}$ to represent the $ith$ input trajectory, where $s$ denotes the number of time points.

The transformer encoder described in \cref{sec:methods} for aggregation over time as well as of the time-agnostic and time-aware attention pooling each use 8 attention heads.

\subsection{Experiment 1}
For pilot experiments, the encoder consists of 4 layers, 16 attention heads and an embedding dimension of 256 whereas the decoder consists of 12 layers, 16 attention heads and an embedding dimension of 512.
We train using the Adam optimizer \citep{kingma2017adammethodstochasticoptimization} with standard hyperparameters and a cosine annealing learning rate schedule. The schedule begins with 1,000 warm-up steps and cycles every 30,000 steps, with a period multiplier of 1.1 and a shrinkage factor of 0.75. The learning rate oscillates between a maximum of $2 \times 10^{-4}$ and a minimum of $1 \times 10^{-9}$. Training is performed with a batch size of 55, corresponding on average to about 135 unique ODE systems and roughly 180,000 numerical tokens per batch.

Inference is carried out using beam sampling with a beam temperature of 0.1 and beam size of 20.

\subsection{Experiment 2}
For secondary experiments, the encoder consists of 4 layers, 16 attention heads and an embedding dimension of 256 whereas the decoder consists of 16 layers, 16 attention heads and an embedding dimension of 512.
We increase the batch size to 100 and switch to the Noam scheduler \citep{vaswani2023attentionneed} with 2,000 warm-up steps, again using Adam with standard hyperparameters. The maximum learning rate is chosen heuristically for stability across settings: $4 \times 10^{-4}$ for 2D models and $1 \times 10^{-4}$ for all 3D models. These choices reflect our empirical observation that higher learning rates accelerate early training but may induce instability or divergence in higher-dimensional systems.  All models are trained for $\sim$1.95M steps with batch size 100. 

Inference again uses beam sampling with temperature 0.1 and beam size of 20.

\xhdr{Inference-time rescaling.} \label{app:sec:rescaling}
We also introduce a rescaling scheme that can be used during inference similar to of \citet{d'ascoli2024odeformer}; however, unlike ODEFormer’s approach, our scheme is adapted to handle multiple instances by enforcing a single shared scaling factor across the entire system. Since each instance $\mathbf{x}_j$ begins with different initial conditions, rescaling them independently would yield incompatible representations and prevent proper inversion of symbolic predictions. Instead, we compute the scaling factor $R$ for the rescaling $\mathbf{\tilde x}_j = \mathbf{x}_j / R$ from the root mean square (RMS) of the initial values across all trajectories,

$$
\frac{1}{R} = \sqrt{\tfrac{1}{N}\sum_{j=1}^N \mathbf{x}_j^2(t_0)},
$$

which robustly captures the overall magnitude of the system without canceling out symmetric values (e.g., +20 and –20). This makes sense because RMS provides a stable measure of the typical scale across instances, ensuring that the decoder receives inputs on a consistent scale. Applying this factor uniformly guarantees compatible representations, enabling unified symbolic predictions that can be reliably mapped back to the original scale.

\section{Details on baseline models} \label{app:sec:baseline_configs}

\xhdr{Using SR models for dynamical symbolic regression.}
Although PySR \citep{cranmer2023interpretablemachinelearningscience} and FFX \citep{McConaghy2011} were originally not proposed to infer differential equations, they can still be applied to this setting with appropriate preprocessing. A straightforward approach is to numerically differentiate observed trajectories to obtain pairs $\{(\dot{\mathbf{x}}_k, \mathbf{x}_k)\}$, which can then be passed to the symbolic regression methods to recover governing equations. SINDy incorporates this step internally, as it is specifically designed for dynamical systems. For PySR and FFX, we implement it explicitly using the \texttt{numpy.gradient} function with default parameters.

However, this procedure is highly sensitive to noise as well as irregular or sparse sampling in time. These limitations motivated our exploration of alternative, more robust formulations of dynamical symbolic regression.

\xhdr{PySR settings.}
For PySR, we configure the symbolic regression search as follows:

\begin{itemize}
    \item Operators:
    \begin{itemize}
        \item Binary: \texttt{+}, \texttt{*}
        \item Unary: \texttt{sin}, \texttt{inv} (inverse), \texttt{pow2} (square)
    \end{itemize}
    This operator set was chosen to align with the dataset and ensure fairness.
    
    \item Loss Function: Mean Squared Error (MSE), computed elementwise.
    
    \item Model Selection Score: Candidate expressions are ranked using PySR’s internal metric that balances accuracy and symbolic complexity, with complexity capped at 30 (default).
    
    \item Computation Constraints: A timeout of 30 seconds per regression task is enforced to provide practical comparability with our model and other baselines. Despite this cap, PySR typically runs slower than other methods due to its exhaustive symbolic search.
    
    \item Search Parameters: Default settings are used for population size, mutation rates, and other hyperparameters. The evolutionary search proceeds until the timeout is reached.
\end{itemize}

\xhdr{SINDy settings.}
We adopt the following configuration for SINDy:

\begin{itemize}
    \item Optimizer: STLSQ
    \item Threshold: 0.1
    \item Max iterations: 400
    \item Differentiation method: finite differences, with order set to 2
    \item Feature library: default (no constraints), allowing exploration of the full set of supported basis functions
\end{itemize}

\xhdr{FFX settings.}
In our experiments, we configured FFX with the following settings:

\footnotesize
\texttt{CONSIDER\_NONLIN = False  \# Disable use of nonlinear functions like abs(), log()} \\
\texttt{CONSIDER\_THRESH = False  \# Disable use of hinge (threshold) functions}
\normalsize

These choices reflect prior knowledge about the underlying ODE systems, which consist of smooth functions. By disabling unnecessary transformations, we constrain FFX to a function space consistent with the data, ensuring a fairer comparison.

\xhdr{ODEFormer settings.} For ODEFormer, we use beam temperature of 0.1 and beam size of 20.

\section{Additional results} \label{app:sec:additional_results}

\begin{figure}[h]
    \centering
    \includegraphics[width=0.24\linewidth]{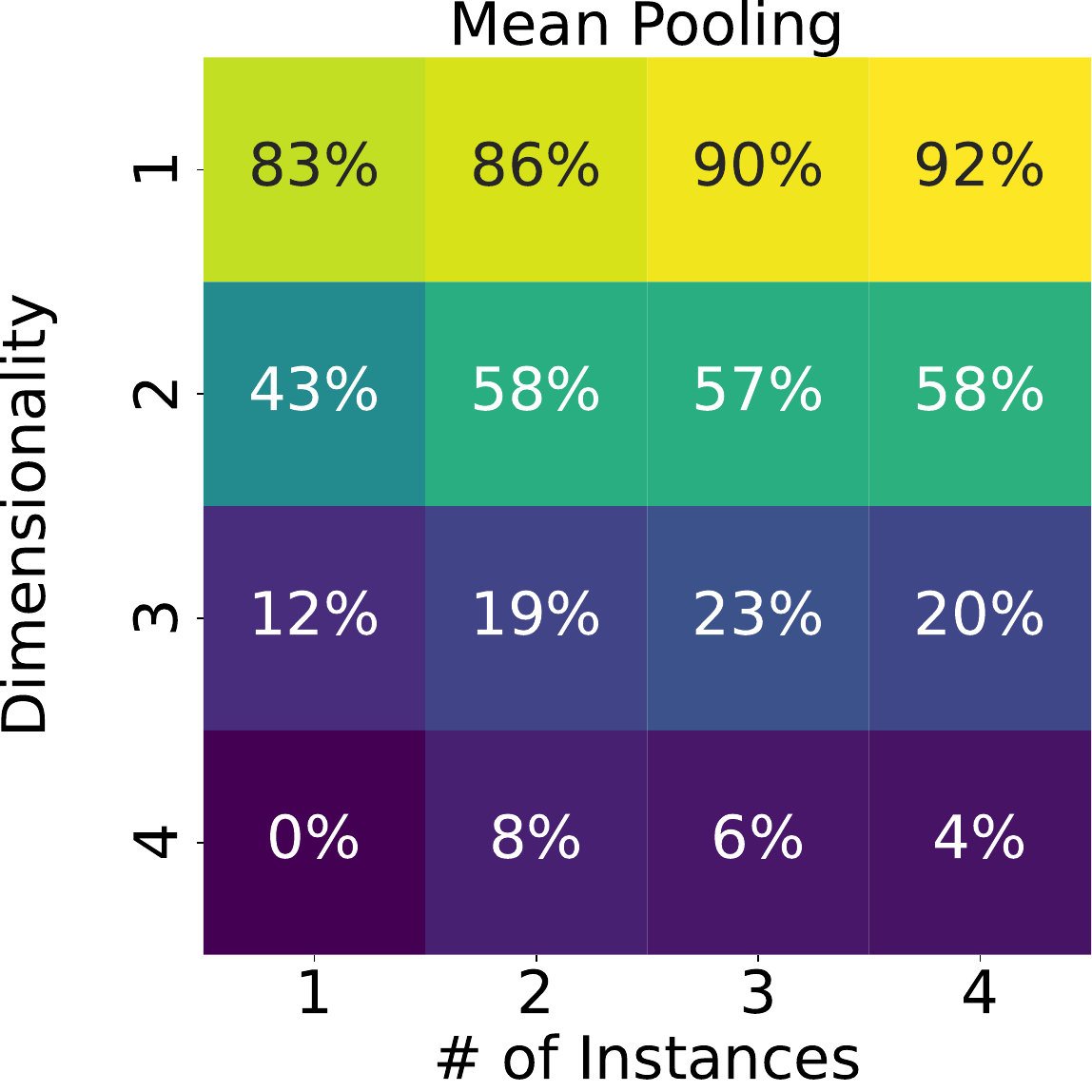}
    \includegraphics[width=0.24\linewidth]{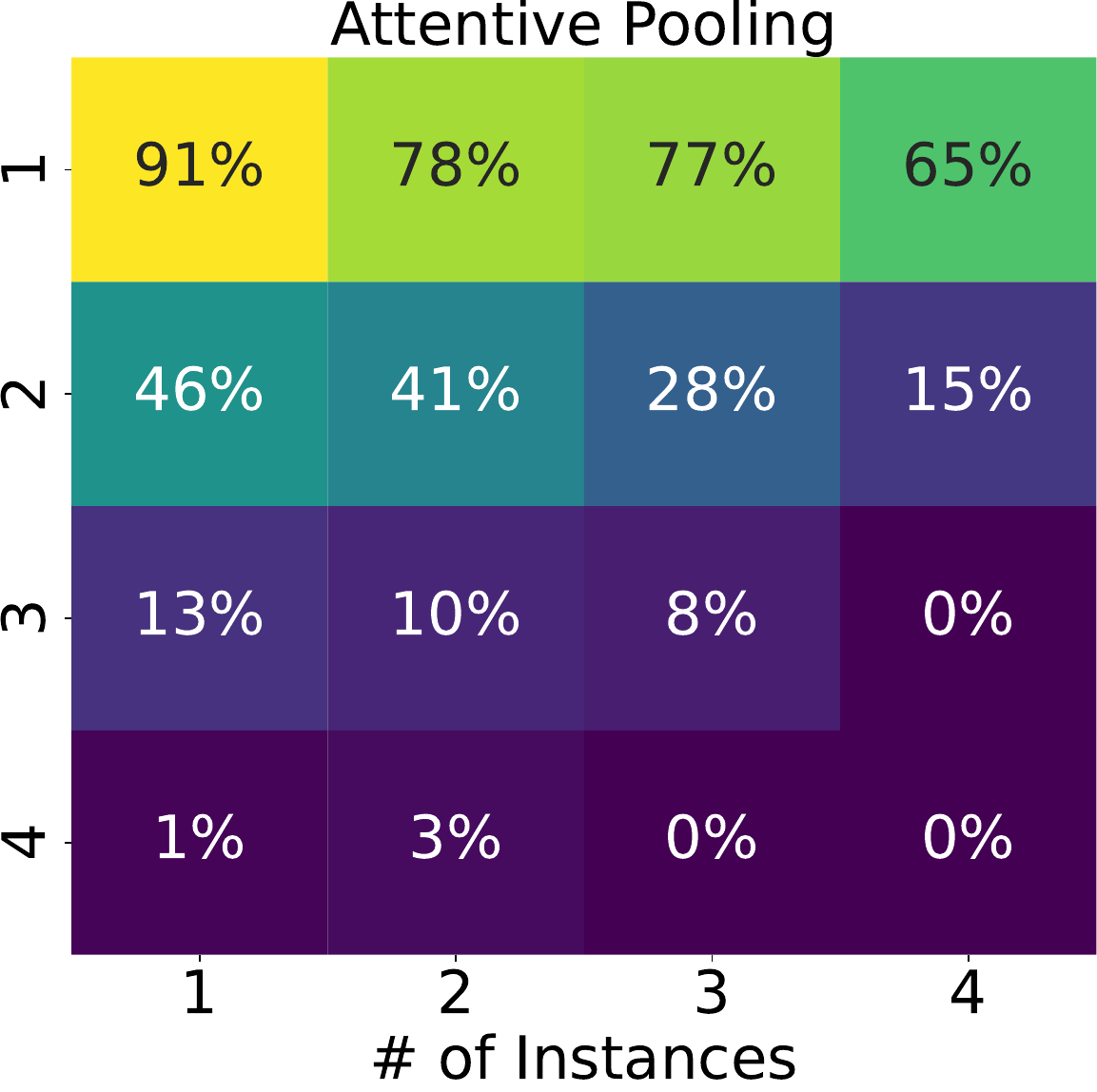}
    \includegraphics[width=0.24\linewidth]{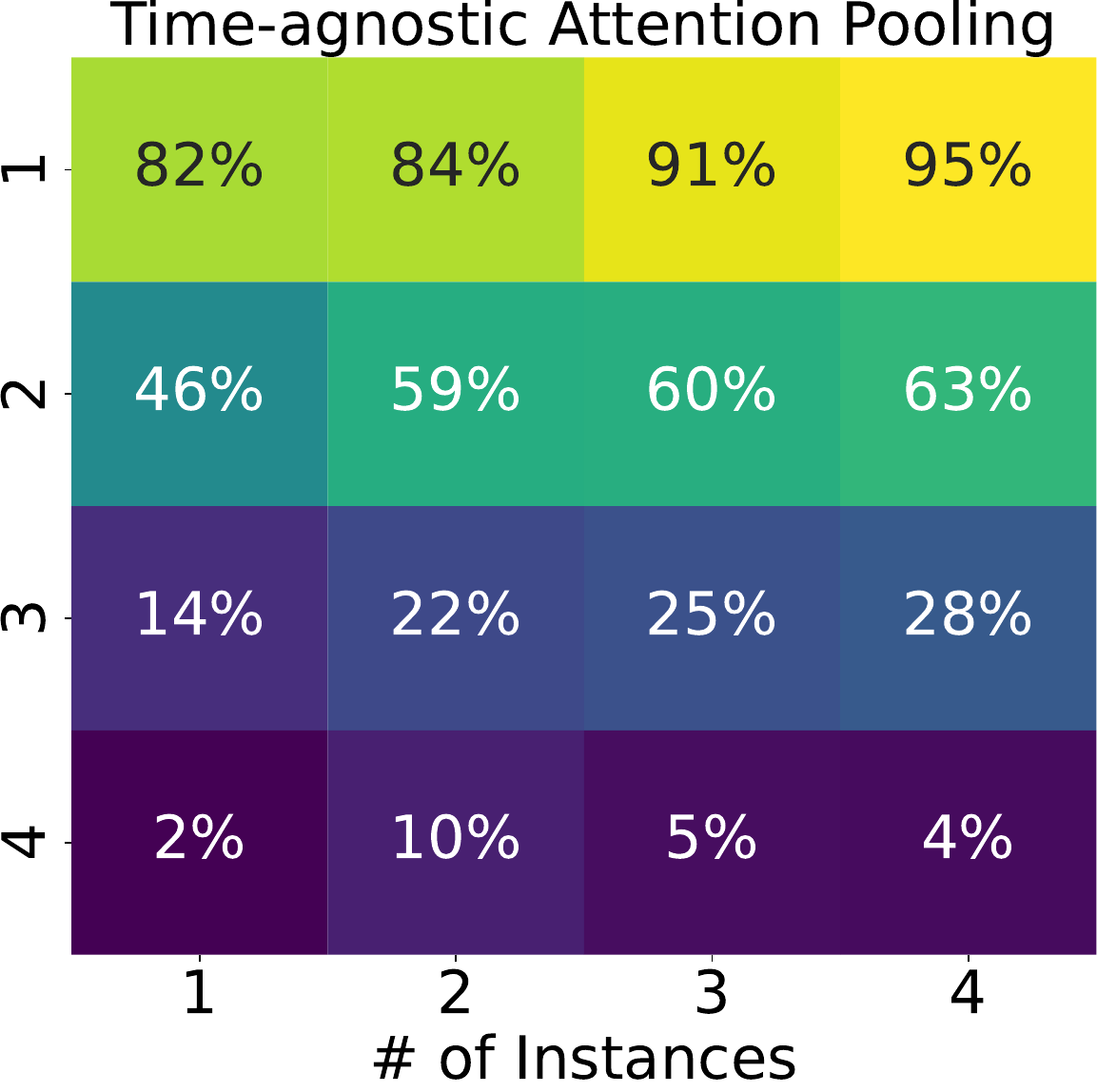}
    \includegraphics[width=0.24\linewidth]{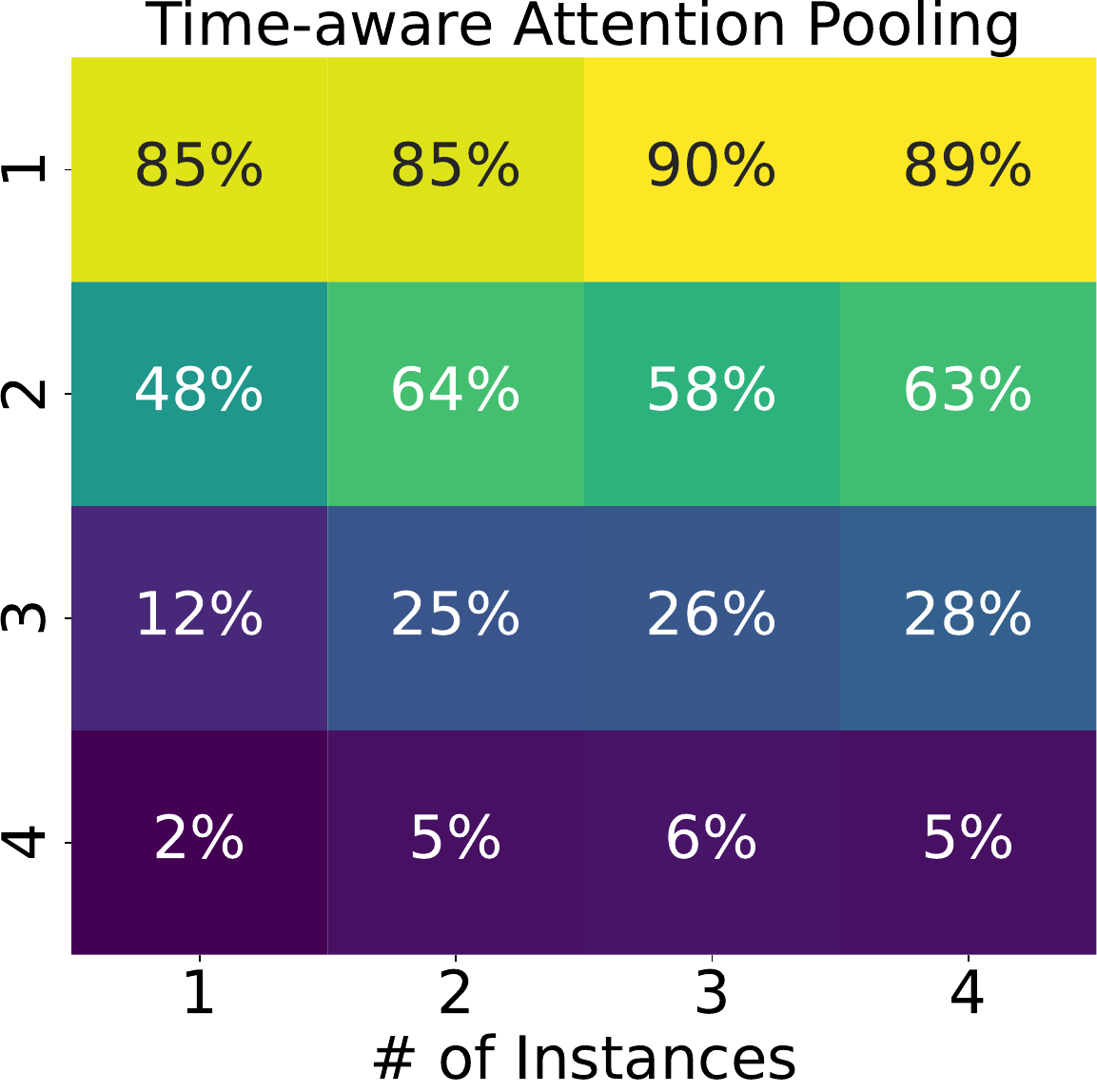}
    \caption{\textbf{Heatmap of different aggregation methods.} We demonstrate how the performance of each aggregation approach depends on number of dimensions and number of instances.}
    \label{app:fig:sec_heatmap}
\end{figure}

\xhdr{Experiment 1: Generalization results resolved in system dimensions and number of instances.} 
We present the generalization results resolved for system dimensions and number of instances in \cref{app:fig:sec_heatmap}. In case of mean pooling, time-agnostic pooling and time-aware pooling we can see that additional instances help in all system dimensions. For attentive pooling this trend is reversed, once again highlighting the inferiority of this aggregation method.

\xhdr{Experiment 2: Reconstruction results.} 

\begin{figure}[h]
    \centering
    \includegraphics[width=0.49\linewidth]{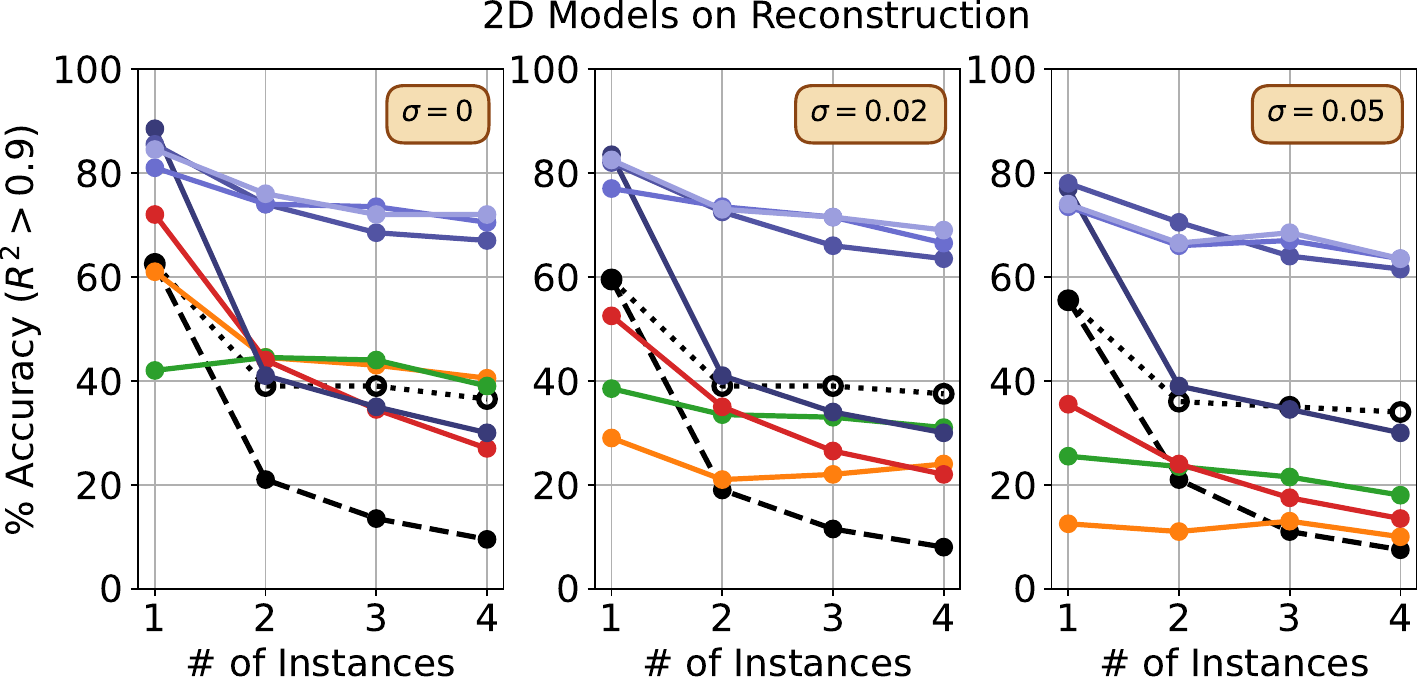}
    \hspace{0.5pt}
    \includegraphics[width=0.49\linewidth]{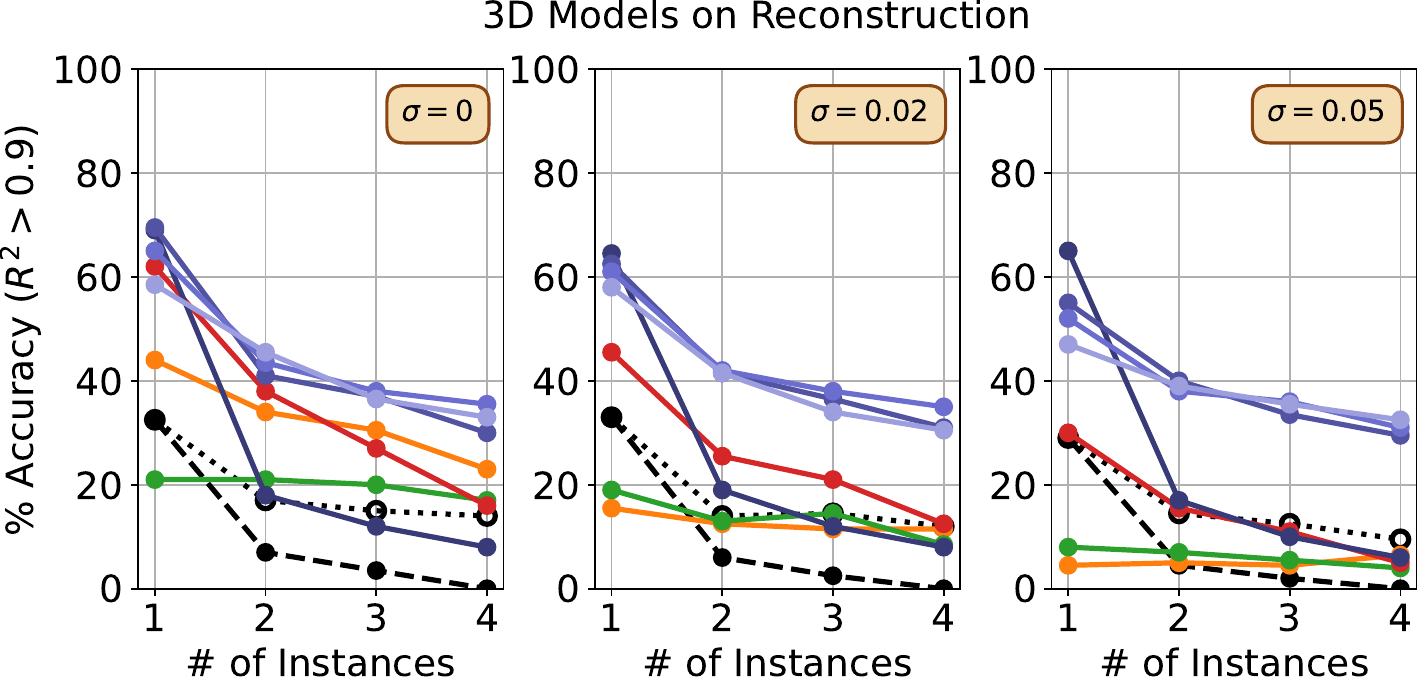}
    \includegraphics[width=\linewidth]{Plots/fig3_legend.pdf}
    \caption{\textbf{Multitraj-ODEFormer and baselines in 2D and 3D scheme.} We compare our models trained with different number of instances with baseline models in reconstruction task.}
    \label{app:fig:pilot_rec}
\end{figure}

The reconstruction results in \cref{app:fig:pilot_rec} reflect the increase in task difficulty as the number of instances increases: although more instances inform the prediction, the ODE also needs to fit more trajectories and in this sense is more constraint. This trend seems more or less universal for all models although MIO-$N$ with $N>1$ is among the least affected models and also shows the best performance across dimensions and noise levels overall.

\end{document}